# Behavior-based Navigation of Mobile Robot in Unknown Environments Using Fuzzy Logic and Multi-Objective Optimization

Thi Thanh Van Nguyen, Manh Duong Phung and Quang Vinh Tran

*University of Engineering and Technology - Vietnam National University, Hanoi*
*vanntt@vnu.edu.vn*

## Abstract

*This study proposes behavior-based navigation architecture, named BBFM, to deal with the problem of navigating the mobile robot in unknown environments in the presence of obstacles and local minimum regions. In the architecture, the complex navigation task is split into principal sub-tasks or behaviors. Each behavior is implemented by a fuzzy controller and executed independently to deal with a specific problem of navigation. The fuzzy controller is modified to contain only the fuzzification and inference procedures so that its output is a membership function representing the behavior's objective. The membership functions of all controllers are then used as the objective functions for a multi-objective optimization process to coordinate all behaviors. The result of this process is an overall control signal, which is Pareto-optimal, used to control the robot. A number of simulations, comparisons, and experiments were conducted. The results show that the proposed architecture outperforms some popular behaviorbased architectures in term of accuracy, smoothness, traveled distance, and time response.*

## 1. Introduction

Mobile robot navigation is one of the most challenging problems in robotics. To complete a navigation task, the robot must be capable of *perceiving* its surrounding environment, *interpreting* data from sensors, *planning* the path to be tracked and *controlling* the actuators to reach the target [1]. Navigation, on the other hand, is fundamental for mobile robot applications. In order to complete any given task, the robot first needs to have the capability of reaching the target safely. Navigation of mobile robot thus has been receiving much research attention and the approaches can be classified into two main categories: *hierarchical architectures* and *reactive* or *behavior-based* architectures [2].

The hierarchical architecture operates through sequent steps of sensing, planning and acting based on a known model of the environment. This architecture is appropriate for static and structured environments. For unknown environments, the behavior-based architecture is often used. This approach splits a complex navigation task into sub-tasks or behaviors as shown in Figure 1. Each behavior is an independent control module dealing with a specific problem of navigation. It takes input data from sensors and generates an output control signal specifying for its objective. The output control signals of all behaviors are then combined in accordance with the global navigating objective to generate an overall control signal. As the combination only uses the local data, the behavior-based architecture does not need to have a global map of the environment. Besides, the use of behaviors enables the modularization and scalability of the architecture.

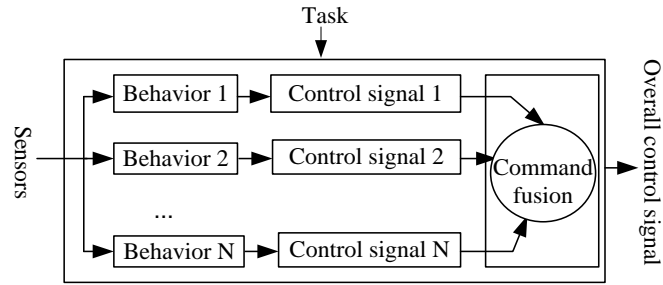

**Figure 1. General Scheme of the Behavior-based Navigation Architecture**

The main challenge with the behavior-based architecture is how to combine behaviors, alled *command fusion*, to achieve the navigation objective efficiently. The *superposition* techniques deal with this problem by using a linear combination of behaviors generated from potential fields [3] or motor schemata [4]. The potential fields select an action based on vector sums of the potential fields produced by an attractive force from goal and repulsive forces from obstacles. The motor schemata use the potential field to define the output of each schema and then combine them to generate a motor action basing on predetermined weighting factors. Those techniques are simple to implement but difficult to adjust gains.

Another command fusion approach is *voting* techniques in which each behavior votes for or against a set of actions. The actions are then combined to generate the best one. In [5], the combination is simply the sum of votes. In DAMN [6], the combination is based on weighting factors assigned by a mode manager. Due to compromise without priority, these techniques may present poor performance in conflict situations, for example, if the "obstacle avoidance" behavior votes for turning right to avoid an obstacle in front of it, while the "goal reaching" behavior votes for turning left since the goal is on the left of it, the compromised action may direct the robot forward resulting a collision with the obstacle.

A command fusion approach commonly used in recent mobile robot navigation systems [7] - [14] is the *fuzzy* technique. This technique presents each behavior by a fuzzy controller. The output fuzzy sets of all controllers are then combined and defuzzified to generate the overall control signal. This technique is simple to implement and quite efficient in navigation. The fusion, however, is not optimal as each defuzzification method often results in a different control value [15], [16].

In order to deal with the optimization problem in command fusion, a technique based on multi-objective optimization theory, called MOASM, was proposed [17]. This technique represents each behavior by an objective function that assigns to each control signal a value reflecting the grade of behavior's objective. A multiobjective optimization process is then applied to find the solution which best maximizes all the objective functions. The main advantage of this technique is its theoretical approach to ensure the optimality of the found solutions. However, the lack of a framework for designing objective functions, which are usually complicated, prevented it from practical use.

In this study, we propose an approach to integrating the advantages of fuzzy logic and multi-objective optimization into a single behavior-based navigation architecture called BBFM. In this architecture, each behavior is represented by a fuzzy controller which only contains the fuzzification and fuzzy inference procedures. Consequently, the output of each fuzzy controller is a function of input variables whose value represents the grade of behavior's objective. They are then used as the input for a multi-objective optimization process to find the optimal value of the overall control signal. The results from a number of simulations, comparisons, and experiments confirm the efficiency of the proposed architecture in navigating the mobile robot in complex and unknown environments.

The structure of paper includes five sections. Section 2 presents the BBFM architecture. Section 3 simulates and compares the BBFM with two other popular architectures. Section 4 presents experimental results. The paper finishes with conclusions in Section 5.

## 2. Proposed Architecture

In this section, the overall structure of the BBFM architecture is first described. After that, detailed implementation of each behavior is introduced. Finally, the command fusion using multi-objective optimization is presented.

### 2.1. Overall Structure of the BBFM

The robot used to evaluate the proposed architecture is the type of differential drive wheeled mobile robot with non-holonomic constraints. Its parameters are shown in Figure 2.

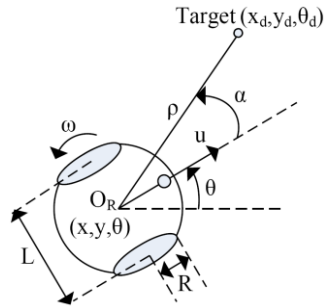

**Figure 2. The Differential Drive Wheeled Mobile Robot and Its Parametes**

where $R$ is the wheel diameter, $L$ is the distance between two wheels, and $(x, y, \theta)$ represents the position and orientation of the robot. Let $(x_d, y_d, \theta_d)$ be the position and orientation of the target. We define three additional variables for navigation: $\rho$ defined by

Equation (1) is the distance from the center of the robot to the target; $\alpha$ defined by Equation (2) is the angle between the robot heading and the vector connecting the robot center with the target; and $e_d$ defined by Equation (3) represents the movement of the robot with the target. For the sake of simplicity, whenever we refer to the position of the robot in this paper, we mean its position and orientation.

$$\rho = \sqrt{(x_d - x)^2 + (y_d - y)^2} \tag{1}$$

$$\alpha = \arctan(y_d - y, x_d - x) - \theta, \alpha \in [-\pi, \pi] \tag{2}$$

$$e_d = \rho_i - \rho_{i-1} \tag{3}$$

In the system, the motion of robot is controlled by adjusting its linear velocity, $u$, and angular velocity, $\omega$. The position of robot is determined via optical quadrature encoders. To sense the environment, the robot is equipped with eight ultrasonic sensors clustered into three groups of left, right, and front as shown in Figure 3. The measuring value of each group is the minimum value of all sensors in that group:

$$\begin{aligned} d_r &= \min(d_1, d_2, d_3) \\ d_f &= \min(d_4, d_5) \\ d_l &= \min(d_6, d_7, d_8) \end{aligned} \tag{4}$$

where $d_i$ is the distance to obstacles measured by sensor $i$. The mission of the robot is to navigate in an unknown environment from an initial position to a desired target without colliding with obstacles and getting trapped in any trap area.

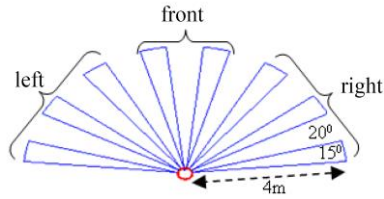

**Figure 3. Arrangement of Ultrasonic Sensors on the Robot**

In order to complete this task, it is natural to subdivide it into small and easy-to-manage behaviors that focus on execution of specific subtasks. The subtasks would be: (i) reaching the target from an arbitrary position, (ii) avoiding obstacles, and (iii) escaping local minimum (trapped) regions. As a result, this approach simplifies the navigation solution while offering a possibility to add new behaviors to the system without causing any major increase in complexity. Individual behavior, however, needs to cope with uncertainties and incompleteness of sensory information as well as with the fact that the operating environment contains elements of dynamics and variability. Fuzzy logic is known to be an organized method for dealing with those problems. Using linguistic rules, fuzzy logic requires neither mathematical models of the environment nor the robot dynamics to design motion controllers. Instead, it fuzzifies the inputs and takes advantage of expert knowledge to discover and represent data relationships and to improve uncertainty reasoning. Therefore, fuzzy logic provides a framework for designing individual behavior.

The command fusion using fuzzy logic, however, does not give reliable solutions. Figure 4 shows that the two ways of fusion using fuzzy logic: defuzzifying first and then combining individual decisions, and combining individual decisions first and then defuzzifying, generate different results. The reason is due to the concurrent activation of multiple behaviors with possibility of conflict between them. In this context, it may not exist a globally optimal solution that is simultaneously optimal with respect to all behaviors. The optimization of one behavior's objective might be associated with a simultaneous deterioration of other behavior's objective. Thus, it is more appropriate to search for a "good enough" solution that can guarantee a suitable trade-off between a multitude of conflicting behaviors' objectives. Multi-objective optimization provides a theoretical approach to find this solution. It quantifies the problem through objective functions that assign to each input a value reflecting their objectives' desirability and provide methods to optimize those functions simultaneously. Interestingly, if we remove the defuzzification procedure in the design process of fuzzy controllers, the output of each behavior will be a fuzzy membership function that its value represents the grade of behavior's objective. This function, therefore, encrypts the semantics meaning of objective functions. This way, the fuzzy technique and multi-objective optimization can be combined.

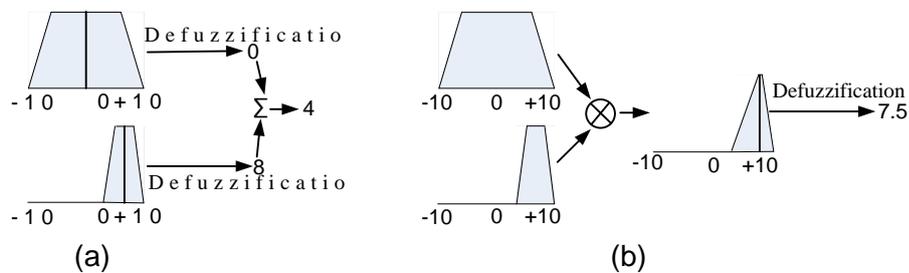

**Figue 4. Two Fuzzy-based Approaches to Command Fusion: (a) Defuzzifying First and Then Combining, (b) Combining First and Then Defuzzifying**

Figure 5 shows the structure of our proposed architecture. It has three behaviors including the local minimum avoidance, obstacle avoidance, and goal reaching. Each behavior is implemented by one customized fuzzy controller consisting of fuzzification and inference modules. The inputs include variables defined in equations (1) - (4). The output of each fuzzy controller includes two membership functions that map the input space to the interval of [0, 1] representing the grade of behavior objective. They are then used as the objective functions for a multi-objective optimization process to generate the overall control signals, $u^*$ and $\omega^*$, which are the Pareto-optimal solutions.

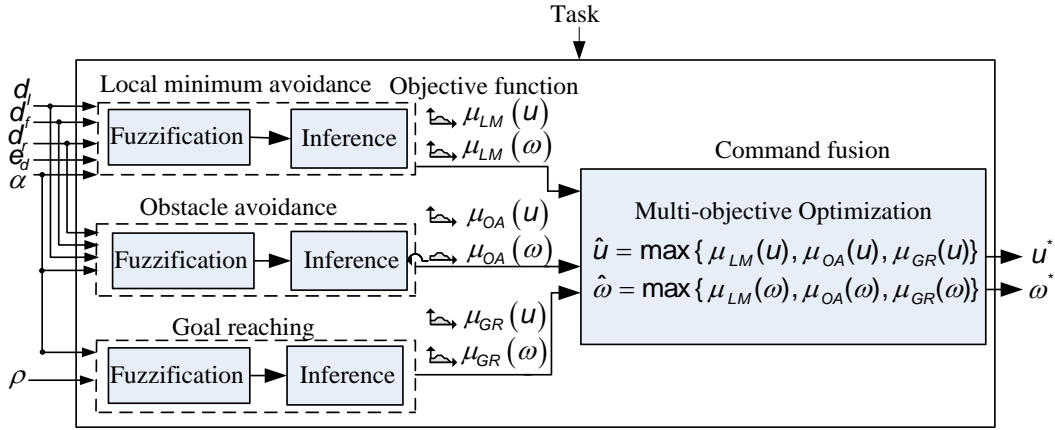

**Figure 5. The Overall Structure of the BBFM**

## 2.2. Design of Individual Behavior

Each behavior is implemented by one customized fuzzy controller which only contains the fuzzification and inference procedures, ignoring the defuzzification. The fuzzification procedure maps the crisp input values to the fuzzy linguistic terms. Each linguistic term is determined by a fuzzy set that is characterized by its membership function. In this study, the Gaussian and Sigmoid membership functions are chosen to represent the fuzzy sets as follows:

$$\mu_{A_{ij}} = e^{\frac{-(x_i - c_{ij})^2}{2(\sigma_{ij})^2}} \quad (5)$$

$$\mu_{B_{lk}} = e^{\frac{-(y_l - c_{lk})^2}{2(\sigma_{lk})^2}} \quad (6)$$

for the input and output Gaussian membership functions, respectively, and

$$\mu_{A_{ij}} = \frac{1}{1 + e^{-a_{ij}(x_i - b_{ij})}} \quad (7)$$

$$\mu_{B_{lk}} = \frac{1}{1 + e^{-a_{lk}(x_l - b_{lk})}} \quad (8)$$

for the input and output Sigmoid membership functions, respectively, where $x_i$ is the $i$th input variable, $y_l$ is the $l$th output variables, $A_{ij}$ is the $j$th fuzzy set of the $i$th input variable, $B_{lk}$ is the $k$th fuzzy set of the $l$th output variable, $\{c_{ij}, \sigma_{ij}\}$ and $\{c_{lk}, \sigma_{lk}\}$ are the parameters of input and output Gaussian membership functions, respectively, and $\{a_{ij}, b_{ij}\}$ and $\{a_{lk}, b_{lk}\}$ are the parameters of input and output Sigmoid membership functions, respectively.

The inference procedure is responsible for formulating the relationship between the inputs and the outputs. It is based on rules of the form" if...then...", for example," if $x_1 = A_{1j}$ and $x_2 = A_{2j}$ and . . . $x_m = A_{mj}$ then $y_1 = B_{1k}$ and $y_2 = B_{2k}$ and . . . $y_n = B_{nk}$". The result of each rule for each output variable is then given by:

$$\mu_{R_k}(y_l) = \min(H, \mu_{B_{lk}}(y_l)),$$
$$H = \min\{\mu_{A_{1j}}(x_1), \mu_{A_{2j}}(x_2), ...., \mu_{A_{mj}}(x_m)\} \quad (9)$$

where $R_k$ denotes the $k$th rule. For $M$ control rules, the implication result of each output variable according to the max-min method is an aggregated output fuzzy set with the membership function determined by:

$$\mu_R(y_l) = \max(\mu_{R_1}(y_l), \mu_{R_2}(y_l), ...., \mu_{R_M}(y_l)). \quad (10)$$

The membership function (10) represents the degree of membership of each element in the set of output variable $y_l$. Detailed implementations of the fuzzification and inference procedures for each behavior are described as follows.

### 2.2.1. Obstacle Avoidance

The obstacle avoidance behavior includes four input variables $d_r$, $d_f$, $d_l$, and $\alpha$, and two output variables $u$ and $\omega$ as shown in Figure 5. Their linguistic terms and membership functions are defined as shown in Figure 6.

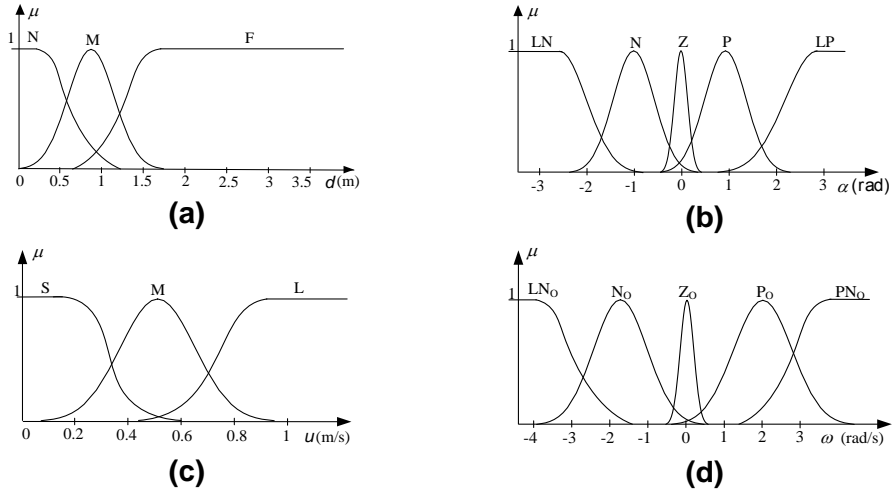

**Figure 6. The Linguistic Terms and Membership Function of Input and Output Variables of the Obstacle Avoidance Behavior: (a) $d_l$, $d_f$, $d_r$; (b) $\alpha$; (c): $u$; (d): $\omega$**

Table 1 presents 28 control rules defined for the behavior. Let $\mu_{R_{OA,k}}(u)$ and $\mu_{R_{OA,k}}(\omega)$ be respectively the results of the $k$th rule in this table for $u$ and $\omega$ by using Equation (9). The implication results according to the max-min method are then given by:

$$\mu_{R_{OA}}(u) = \max(\mu_{R_{OA,1}}(u), \mu_{R_{OA,2}}(u), ..., \mu_{R_{OA,28}}(u))$$
$$\mu_{R_{OA}}(\omega) = \max(\mu_{R_{OA,1}}(\omega), \mu_{R_{OA,2}}(\omega), ..., \mu_{R_{OA,28}}(\omega)). \quad (11)$$

**Table 1. Rules for Obstacle Avoidance**

| Collisions | Rule | Input | | | | Output | |
|---|---|---|---|---|---|---|---|
| | | $d_l$ | $d_f$ | $d_r$ | $\alpha$ | $u$ | $\omega$ |
| | 1 | N | N | F | | S | Po |
| | 2 | F | N | N | | M | Po |
| | 3 | M | N | N | | M | Po |
| | 4 | F | N | M | | N | LPo |
| | 5 | N | N | F | | M | LNo |

| | 6 | N | N | M | | M | No |
|---|---|---|---|---|---|---|---|
| | 7 | F | N | F | | M | Lpo |
| | 8 | M | N | F | | M | No |
| | 9 | F | N | M | | M | Po |
| | 10 | N | M | N | | M | Po |
| | 11 | N | N | N | | S | Po |
| | 12 | M | N | M | | S | Po |
| | 13 | N | M | M | | M | No |
| | 14 | N | M | F | | M | No |
| | 15 | N | F | M | | M | No |
| | 16 | N | F | F | LN | S | Lno |
| | 17 | N | F | F | N | S | No |
| | 18 | N | F | F | Z | L | Zo |
| | 19 | N | N | N | LP | L | Zo |
| | 20 | N | F | F | P | L | Zo |
| | 21 | M | M | N | | M | Po |
| | 22 | F | M | N | | M | Po |
| | 23 | M | F | N | | M | Po |
| | 24 | F | F | N | LN | L | Zo |
| | 25 | F | F | N | N | L | Zo |
| | 26 | F | F | N | Z | L | Zo |
| | 27 | F | F | N | LP | S | Lpo |
| | 28 | F | F | N | P | S | LPo |

**2.2.2. Goal Reaching**

The objective of goal-reaching behavior is to control the robot to reach the target in the shortest time. In order to complete this task, the behavior continuously adjusts the robot direction to match the target direction while driving the robot at the fastest possible speed. The behavior uses two input and two output variables as shown in Figure 5. Three of them including the deflection angle α and the velocities $u$ and $\omega$ have the same definition of linguistic terms and membership functions as in the obstacle avoidance behavior. The fourth variable, $\rho$, has the linguistic terms and membership functions defined as shown in Figure 7.

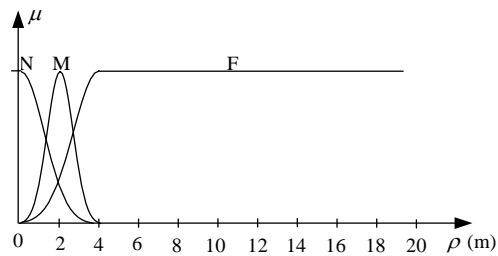

**Figure 7. The Linguistic Terms and Membership Function of ρ**

The behavior has 15 rules defined as in Table 2. Let $\mu_{R_{GR,k}}(u)$ and $\mu_{R_{GR,k}}(\omega)$ be respectively the results of the $k$th rule in the table for output variables $u$ and $\omega$ by using Equation (9). The implication results for $u$ and $\omega$ according to the max-min method are then given by:

$$\mu_{R_{GR}}(u) = \max(\mu_{R_{GR,1}}(u), \mu_{R_{GR,2}}(u), ..., \mu_{R_{GR,15}}(u))$$
$$\mu_{R_{GR}}(\omega) = \max(\mu_{R_{GR,1}}(\omega), \mu_{R_{GR,2}}(\omega), ..., \mu_{R_{GR,15}}(\omega)). \tag{12}$$

**Table 2. Rules for Goal Reaching**

| Rule | Input | | Output | |
|---|---|---|---|---|
| | $\rho$ | $\alpha$ | $u$ | $\omega$ |
| 1 | N | Z | S | Zo |
| 2 | N | N | S | No |
| 3 | N | LN | S | LNo |
| 4 | N | P | S | Po |
| 5 | N | LP | S | LPo |
| 6 | M | Z | M | Zo |
| 7 | M | N | M | No |
| 8 | M | LN | M | LNo |
| 9 | M | P | M | Po |
| 10 | M | LP | M | LPo |
| 11 | F | Z | L | Zo |
| 12 | F | N | L | No |
| 13 | F | LN | L | LNo |
| 14 | F | P | L | Po |
| 15 | F | LP | L | LPo |

**2.2.3. Local Minimun Avoidance**

During navigation, the robot may be trapped in U-shape regions called local minimum problem. It happens when the robot repeatedly conducts opposite turning commands during avoiding obstacles. Figure 8(a) illustrates such a situation. The robot starts from A, moves toward the target according to rules 6 and 11 in Table 1; at B, it turns left according to rules 11 and 12 to move to C or it can turn right to move to D depending on the priority; at C, it turns left to go toward the target according to rules 27 and 28 and back to B. The process is repeated causing the robot to be trapped.

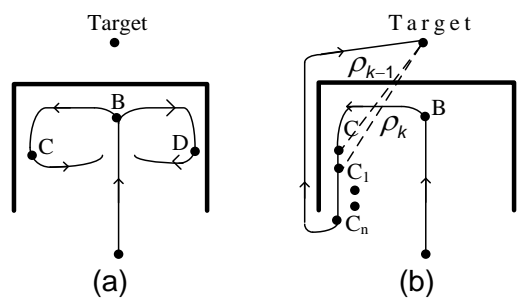

**Figure 8. The Fuzzy-based Method for Dealing with the Local Minimum Problem: (a) the Robot is trapped in a Local Minimum Region; (b) the Robot Scapes the Local Minimum Region Using the Fuzzy-based Method**

In order to deal with this problem, we carefully analyze various trapped situations and realize that it is critical to detect the trapped points which are points C and D in the example shown above. We detect them by checking if the distance $e_d$ is increasing, the target is behind the robot (corresponding to the positive value of $\alpha$), and the obstacle is on the right (or left) side of robot. Figure 8(b) shows that if the robot goes straight instead of turning left at the trapped point C, it will reach $C_n$ as the conditions do not change from C to $C_n$. At $C_n$, the robot will turn right to exit the local minimum region. In order to implement the proposed solution, the behavior of local minimum avoidance has five input variables including $d_l$, $d_f$, $d_r$, $\alpha$, and $e_d$ and two output variables including $u$ and $\omega$. The linguistic terms and membership functions of $e_d$ are defined as in Figure 9.

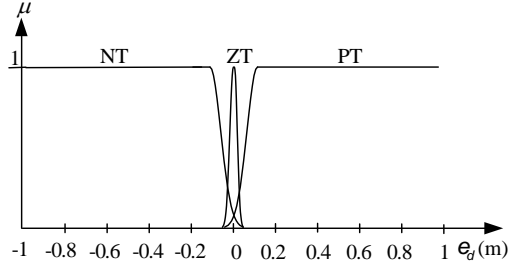

**Figure 9. The Linguistic Terms and Membership Function of $e_d$**

The linguistic terms and membership functions of the remained variables are defined as in the obstacle avoidance behavior. Table 3 presents 7 control rules for detecting trapped points and escaping local minimum regions. Let $\mu_{R_{LM,k}}(u)$ and $\mu_{R_{LM,k}}(\omega)$ be respectively the results of the $k$th rule in the table for output variables $u$ and $\omega$ by using Equation (9). The implication results for $u$ and $\omega$ according to the max-min method are then given by:

$$\mu_{R_{LM}}(u) = \max(\mu_{R_{LM,1}}(u), \mu_{R_{LM,2}}(u), ..., \mu_{R_{LM,7}}(u))$$
$$\mu_{R_{LM}}(\omega) = \max(\mu_{R_{LM,1}}(\omega), \mu_{R_{LM,2}}(\omega), ..., \mu_{R_{LM,7}}(\omega)). \quad (13)$$

**Table 3. Rules for Local Minimum Avoidance**

| Rule | Input | | | | | Output | |
|---|---|---|---|---|---|---|---|
| | $d_l$ | $d_f$ | $d_r$ | $e_d$ | α | $u$ | $\omega$ |
| 1 | N | N | N | | Z | S | Po |
| 2 | F | N | N | | P | M | Po |
| 3 | F | N | N | | LP | S | LPo |
| 4 | F | F | N | PT | P | M | Zo |
| 5 | F | F | N | PT | LP | M | Zo |
| 6 | F | F | F | PT | P | M | Zo |
| 7 | F | F | F | PT | LP | M | LNo |

### 2.3. Command Fusion

The command fusion is carried out by using multiobjective optimization. As shown in Figure 5, the membership functions of output variables are used as the objective functions. Let $\mu_i(y)$ be the $i$th objective function, $y$ be an output control signal ($y = u$ or $y = \omega$), $Y$ be the domain of $y$, and $N$ be the number of objective functions. The optimal value of each output control signal is then the solution of the following equation:

$$\hat{y} = \arg\max[\mu_1(y), \mu_2(y), ..., \mu_N(y)] \quad (14)$$

According to the theory of multi-objective optimization, there might not exist the optimal solution, $\hat{y}$, of Equation (14), but only the *"good enough"* solution, $y^*$, which is the best fit for all objectives. This solution is called the Pareto-optimal solution or non-dominated solution defined as follows: $y^*$ is the Pareto-optimal solution of Equation (14) if there does not exist any $y \in Y$ such that $\mu_i(y) > \mu_i(y^*)$ for at least one $i$ and $\mu_j(y) \geq \mu_j(y^*)$ for all $j$. In other words, the Pareto optimal solution is the one in which there is not other solution that improves an objective without resulting in the deterioration of at least another objective.

In order to find the Pareto-optimal solution, there exist a number of methods [17] such as weighting, lexicographic, and goal programming. In the context of command fusion, we choose to use the lexicographic method due to its efficiency. This method first requires ranking the order of importance of all objectives, for instance, $\mu_1$ is the most important and $\mu_N$ is the least important. The Pareto-optimal solutions are then obtained by solving the following sequence of problems until either a unique solution is found or all the problems are solved:

$$P_1 : \max_{y \in Y}[\mu_1(y)]$$
$$P_2 : \max_{y \in Y_1}[\mu_2(y)]$$
$$...$$
$$P_N : \max_{y \in Y_{N-1}}[\mu_N(y)]$$
$$Y_i = \{y \mid y \text{ solves } P_i\}, i = 1,..., N-1$$
(15)

In our system, the optimal values of the overall control signal are determined by the output membership functions of (11), (12), and (13) as follows:

$$\hat{u} = \arg\max[\mu_{R_{OA}}(u), \mu_{R_{GR}}(u), \mu_{R_{LM}}(u)],$$
$$\hat{\omega} = \arg\max[\mu_{R_{OA}}(\omega), \mu_{R_{GR}}(\omega), \mu_{R_{LM}}(\omega)]$$
(16)

The lexicographic method used to find the Pareto optimal solutions of (16) are carried out as follows:

- Sorting all behaviors in descending order of importance: local minimum avoidance, obstacle avoidance, and goal reaching.
- Sequentially solving equations $P_i$ by using discrete values of $u$ and $\omega$ on their domains $U$ and $W$ until a unique solution is obtained, or all equations are solved:

$$u^* : \begin{cases} P_1 : \max_{u \in U}[\mu_{R_{LM}}(u)], \\ P_2 : \max_{u \in U_1}[\mu_{R_{OA}}(u)], \\ U_1 = \{u \mid u \text{ solves } P_1\}, \\ P_3 : \max_{u \in U_2}[\mu_{R_{GR}}(u)], \\ U_2 = \{u \mid u \text{ solves } P_2\} \end{cases} \quad \omega^* : \begin{cases} P_1 : \max_{\omega \in W}[\mu_{R_{LM}}(\omega)], \\ P_2 : \max_{\omega \in W_1}[\mu_{R_{OA}}(\omega)], \\ W_1 = \{\omega \mid \omega \text{ solves } P_1\}, \\ P_3 : \max_{\omega \in W_2}[\mu_{R_{GR}}(\omega)], \\ W_2 = \{\omega \mid \omega \text{ solves } P_2\} \end{cases}$$
(17)

- If more than one Pareto-optimal solution is obtained, the one with the greatest value of $u$ and the smallest value of $\omega$ is chosen.

## 3. Simulations

Simulations have been conducted to evaluate the efficiency of the BBFM compared to two other popular architectures including the MOASM [17] and CDB [10]. The MOASM uses multi-objective optimization for command fusion. It is implemented with three behaviors including avoiding obstacles, maintaining the target heading and moving fast forward. The objective functions are defined as in the origin [17]. The overall control signal is determined by using the lexicographic method. The CDB is implemented with three behaviors as in the MOASM. Each behavior is a fuzzy controller. The overall control signal is determined by using fuzzy meta rules and defuzzification. In order to ensure the fairness in comparison, the BBFM only uses the obstacle avoidance and goal reaching behaviors.

All architectures are simulated in Matlab and use the same robot configuration. Its mechanical parameters are set as follows: $R = 0.085$ m, $L = 0.265$ m, $u \in [0, 1.3]$ m/s, and $\omega \in [-4.3, 4.3]$ rad/s. The ultrasonic sensors have the sensing range from 0 m to 4 m and the radiation cone of $15°$. They are arranged in front of the robot as shown in Figure 3 to cover the range of $160°$. The universe of discourse of $\rho$ is in the range of $[0, 20]$ and that of $e_d$ is $[-1, 1]$. In simulations, three scenarios were chosen for navigating the robot. In each scenario, 15 runs were carried out to evaluate the performance of each architecture. Details and results of each scenario are presented as follows.

## 3.1. Popular Operating Environment

In scenario 1, the operating environment is chosen to be the same as in the original paper of MOASM [17]. The start position is (-2, -1.8, 180°) and the target position is (-6, -4.8, 0°). Figure 10 shows the path of robot generated by three architectures MOASM, BBFM, and CDB. It can be seen that all architectures successfully navigate the robot to avoid obstacles and reach the target. Table 4 shows the average performance of each architecture, where the index smoothness is the average absolute value of the difference between the current and the previous orientation, thus showing how smooth the maneuvers is; the index target error is the distance between the final position of the robot and the target position, thus evaluating the reachable ability to the target at the steady state; and the indexes traveled distance and elapsed time are respectively the total distance of the robot's traveled path and the time taken to go through that path. As can be inferred from Table 4, the BBFM is more efficient than the remaining architectures in almost all criteria.

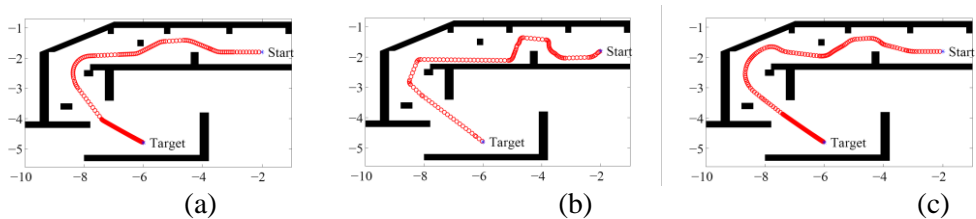

(a)  (b)  (c)

**Figure 10. Travelling Paths of the Robot Navigated by Three Architectures in Scenario 1: (a) BBFM, (b) MOASM, (c) CDB**

**Table 4. Navigation Results in Scenario 1**

| Index | BBFM | MOASM | CDB |
|---|---|---|---|
| Traveled distance (m) | 10.36 | 11.02 | 11.02 |
| Elapsed time (second) | 28.26 | 41.45 | 36.43 |
| Smoothness (degree) | 0.88 | 1.29 | 6.1 |
| Target error (m) | 0.05 | 0.2 | 0.05 |

## 3.2. Office-like Operating Environment

In scenario 2, the operating environment is chosen to be more like an office with wall and bulkhead obstacles. The start position is (-7, -6, 0°) and the target is (-2.5, -1.5, 0°). Figure 11 and Table 5 show the navigation results. It can be seen that the MOASM does not complete the navigation task as its objective functions are built based on the principle of Instantaneous Center of Curvature (ICC) of differential drive wheeled mobile robot which does not efficient in escaping corners. On the other hand, both the BBFM and CDB can safely navigate the robot to reach the target due to the efficiency of fuzzy control. However, the BBFM is more efficient than the CDB.

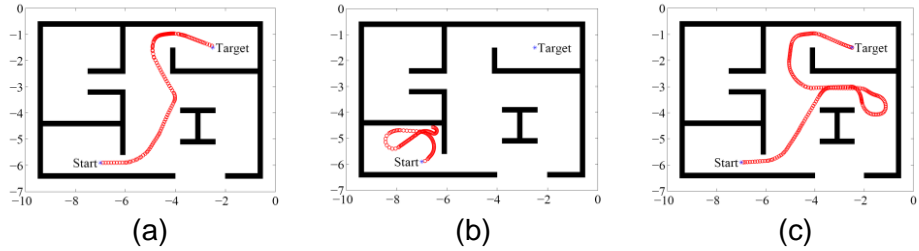

(a)                  (b)                  (c)

**Figure 11. Travelling Paths of the Robot Navigated by Three Architectures in Scenario 2: (a) BBFM, (b) MOASM, (c) CDB**

**Table 5. Navigation Results in Scenario 2**

| Index | BBFM | CDB |
|---|---|---|
| Traveled distance (m) | 9.35 | 15.66 |
| Elapsed time (second) | 12.09 | 24.45 |
| Smoothness (degree) | 2.04 | 3.16 |
| Target error (m) | 0.05 | 0.05 |

### 3.3. Operating Environment with Local Minimun Regions

In the third scenario, the operating environment has local minimum regions. The robot starts at position (-6, -6, $0^\circ$) and desires to reach position (-2.5, -1.5, $0^\circ$). The results in Figure 12 show that all three architectures are not able to complete the navigation task because of local minimum problem. This problem can be solved if adding a local minimum avoidance behavior to the BBFM as shown in Figure 13(a). Even in a more complicated local minimum situation, the BBFM still completes the navigation task as shown in Figure 13(b).

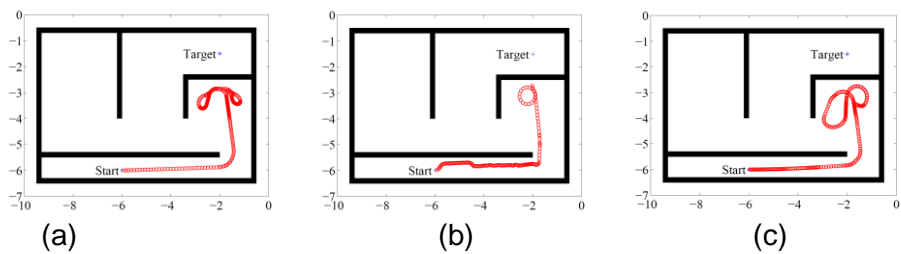

(a)                  (b)                  (c)

**Figure 12. Travelling Paths of the Robot Navigated by Three Architectures in Scenario 3: (a) BBFM, (b) MOASM, (c) CDB**

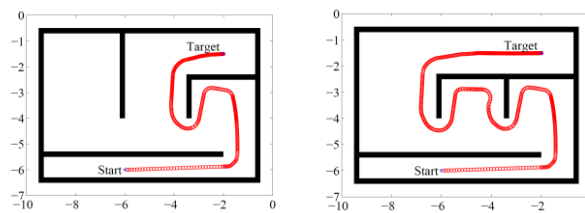

**Figure 13. Travelling Paths of the Robot Navigated by the BBFM in Operating Environment with Local Minimum Regions**

## 4. Experiments

In order to evaluate the performance of the BBFM in real environments, we carried out a number of experiments described as follows.

### 4.1. Experimental Setup

The robot used in experiments is a Sputnik robot [18] as shown in Figure 14. It is equipped with three ultrasonic sensors DUR5200 at left, front and right directions creating the scanning range from $-60°$ to $60°$. To extend the scanning range from $-90°$ to $90°$, we added two additional ultrasonic sensors SRF05 to the left and right sides of the robot. Each employs a microcontroller PIC12F1572 to synchronize its data with the main board of Sputnik robot. The linear and angular velocities of the robot are respectively set to [0, 0.5] m/s and [-3.7, 3.7] rad/s. The position of the robot is determined via optical encoder sensors. The robot has a wireless module connecting it to a Wifi router (Figure 14). The BBFM is written in Matlab and installed on a PC which communicates with the robot through a Wifi router. The sampling time $T_s$ is 300 ms and the experimental environment is an indoor office with the size of 4 m x 3 m and changeable obstacles.

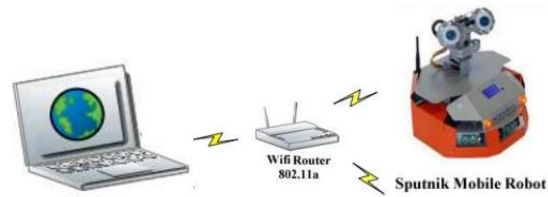

**Figure 14. The Sputnik Robot and Its Configuration to Communicate with the Control Computer**

### 4.2. Experimental Results

Experiments were carried out with different configurations of the environment and target position. Figure 15 shows results of the traveled paths, velocity responses and images of the robot in three of such configurations. In configuration 1 (Figure 15(a)), the robot starts at A (0, 0, 90°) and then moves forward along two walls to B. At B, it turns right two times to C and then goes straight to D. At D, the robot continuously adjusts its direction to avoid the bulkhead corners while on average maintaining the target heading to reach E and finally goes straight to reach the target F (1.9, 2.5, 0°). Figure 15(b) shows the correspondence of linear and angular velocities of the robot with those movements, for instance, between D and E near the target, the angular velocity switches between the left and right directions to avoid the bulkhead corners while the linear velocity gradually decreases to prepare for stopping at the target.

In configuration 2 as shown in Figure 15(d), the environment structure changed with more potential local minimum regions. The robot starts at position (-0.1, -1, 90°) and the target is set to (-0.1, 2.5, 0°). The path from B to D shows that the robot successfully escapes the local minimum region near the starting position. From D to F, it succeeds in avoiding obstacles located at the center and near the target. Figure 15(e) shows the correspondence of the robot's velocities with its movements. The operating environment in Configuration 3 is similar to the configuration 2. However, the start and target positions are changed to (0.1, -0.2, 0°) and (1.8, 2.3, 0°), respectively. As can be inferred from Figure 15(g)-(i), the robot can reach the target while avoiding obstacles and potential local minimum regions.

Table 6 shows the navigation performance in each configuration. As can be inferred from the table, configuration 1 introduces the best performance in smoothness and elapsed time because it is the simplest case. On the contrary, configuration 2 introduces the worst performance due to obstacles and local minimum regions. Nevertheless, the closeness between values of average linear velocities in all configurations implies that the operation of robot is stable and suitable for the indoor environment.

**Table 6. Navigation Results in Three Configurations**

| Configuration | $u_{average}$ (m/s) | Smoothness ( degree) | Elapsed time (s) | Traveled distance (m) | Target error (m) |
|---|---|---|---|---|---|
| 1 | 0.18 | 3.72 | 22.5 | 4.06 | 0.05 |
| 2 | 0.15 | 6.54 | 33 | 4.86 | 0.08 |
| 3 | 0.16 | 5.75 | 24 | 3.77 | 0.07 |

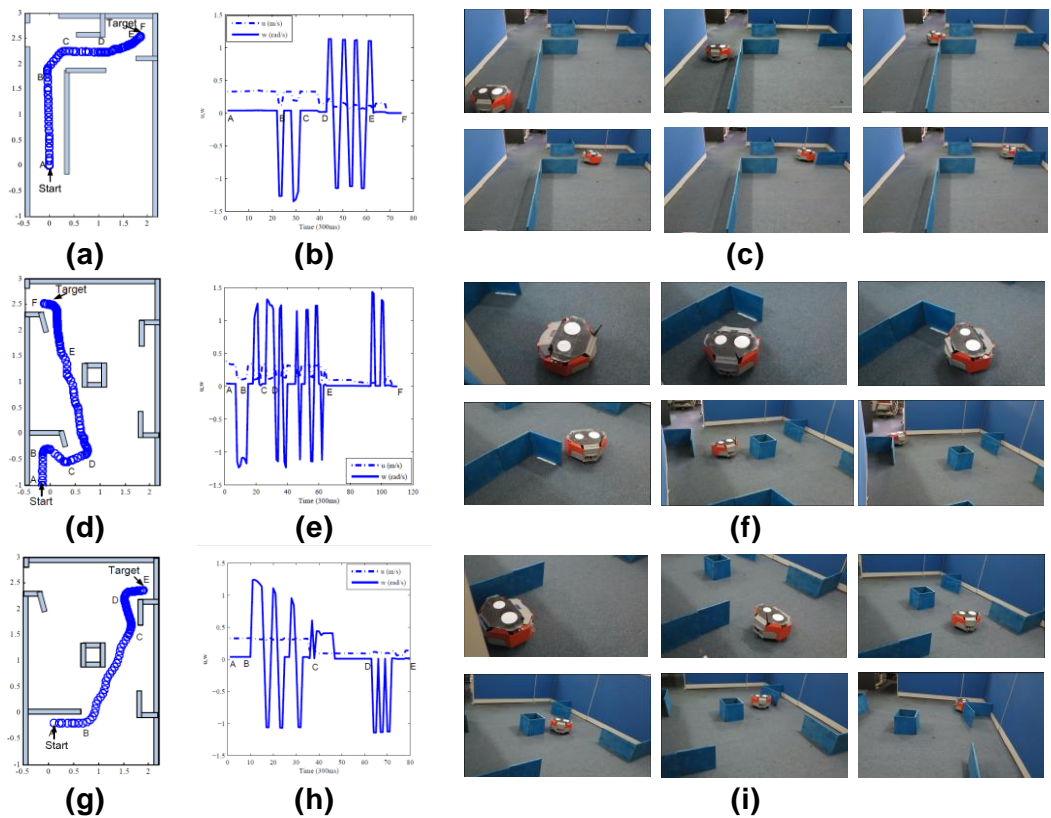

**Figure 15. Travelling Paths, Velocity Responses, and Photos of the Robot Operating in Three Different Navigation Configurations: (a) – (c): Configuration 1, (d) – (f): Configuration 2, (g) – (i): Configuration 3**

## 5. Conclusions

In this paper, we have proposed new behaviorbased navigation architecture for navigating the mobile robot in unknown environments. We modified the procedures of designing fuzzy controllers for behaviors so that their outputs can be used as inputs for a multiobjective optimization process to coordinate the behaviors. Consequently, the architecture inherits advantages of fuzzy logic in dealing with uncertainties of sensory information while providing a framework for designing objective functions. It also takes advantage of multiobjective optimization to generate Pareto-optimal solutions for command fusion. During the development, we proposed a very simple yet effective fuzzy-based approach for dealing with the local minimum problem which often happens in mobile robot navigation. The results show that the proposed architecture is practical in implementation and possibly navigates the robot to reach the target along a smooth and efficient trajectory in environments with unpredictable obstacles, topographies and local minimum regions.

**References**

[1]  S. Roland and N. I. R, "Introduction to autonomous mobile robots", The MIT Press Cambridge, **(2004)**, pp. 10–11.
[2]  D. Nakhaeinia, S. H. Tang, S. B. M. Noor and O. Motlagh, "A review of control architectures for autonomous navigation of mobile robot", International Journal of the Physical Sciences, vol. 6, no. 2, **(2011)**, pp. 169–174.
[3]  O. Khatib, "Real-time obstacle avoidance for manipulators and mobile robots", The International Journal of Robotics Research, vol. 1, no. 5, **(1986)**, pp. 90–98.
[4]  R. Arkin, "Motor-schema based mobile robot navigation", Int. J. Robot, vol. 8, no. 4, **(1989)**.
[5]  J. Hoff and G. Bekey, "An architecture for behavior coordination learning", In IEEE International Conference on Neural Networks, **(1995)**.
[6]  J. Rosenblatt, "DAMN: A distributed architecture for mobile navigation", In Proceedings of the AAAI Spring Symposium on Software Architecture for Physical Agents, **(1995)**.
[7]  Saffiotti, "The uses of fuzzy logic in autonomous robot navigation", Soft Computing, Springer Verlag, **(1997)**, pp. 180–197.
[8]  M. S.-F. Eduardo Freire, Teodiano Bastos-Filho and R. Carelli, "A new mobile robot control approach via fusion of control signals", IEEE transactions on system, mam and cybernetics, vol. 34, no. 1, **(2004)**.
[9]  S. A. Yahmedi, A. B. El-Tahir El-Dirdiri and T. Pervez, "Behavior based control of a robotic based navigation aid for the blind", Control and Applications Conference, **(2009)**.
[10] S. A. Yahmedi and M. A. Fatmi, "Fuzzy logic based navigation of mobile robot", Recent Advances in Mobile Robotics, **(2011)**.
[11] G. Z. Ye and D.-K. Kang, "Fusion of hierarchical behavior-based actions in mobile robot using fuzzy logic", Journal of Information and Communication Convergence Engineering, vol. 10, no. 2, **(2012)**, pp. 149–155.
[12] Q. T. Hongwei Mo and L. Meng, "Behavior - based fuzzy control for mobile robot navigation", Mathematical Problems in Engineering, no. Article ID 561451, **(2013)**.
[13] M. Faisal, K. Al-Mutib, R. Hedjar, H. Mathkour, M. Alsulaiman and E. Mattar, "Behavior based mobile for mobile robots navigation and obstacle avoidance", International Journal of computers and communications, vol. 8, **(2014)**.
[14] R. V. D. Sousa, R. A. Tabile, R. Y. Inamasu and A. J. V. Porto, "A row crop following behavior based on primitive fuzzy behaviors for navigation system of agricultural robots", 4th IFAC Conference on Modelling and Control in Agriculture, Horticulture and Post Harvest Industry, vol. 46, **(2013)**, pp. 91–96.
[15] D. Driankov, H. Hellendoorn and M. Reinfrank, "An introduction to fuzzy control," Spinger, **(2010)**, pp. 132–141.
[16] S. Naaz, A. Alam and R. Biswas, "Effect of different defuzzification methods in a fuzzy based load balancing application", International Journal of Computer Science Issues, vol. 8, no. 1, **(2011)**, pp. 261–267.
[17] P. Pirjanian, "Multiple objective behavior-based control", Robotics and Autonomous Systems, Elsevier, no. 31, **(2000)**, pp. 53–60.
[18] D. R. Manual, http://www.drrobot.com/products.